\documentclass[preprint,12pt,authoryear]{elsarticle}

\usepackage{amssymb}
\usepackage{amsmath,amssymb}
\usepackage{graphicx}
\usepackage{hyperref}
\usepackage{booktabs}
\usepackage{float}
\usepackage{caption}
\usepackage{url}
\usepackage{enumitem} 
\usepackage{threeparttable} 

\journal{Journal of Biomedical Informatics}

\begin{document}

\begin{frontmatter}

\title{Clinical Semantic Intelligence (CSI): Emulating the Cognitive Framework of the Expert Clinician for Comprehensive Oral Disease Diagnosis}

\author[1]{Mohammad Mashayekhi}
\author[2,3]{Sara Ahmadi Majd}
\author[4]{Arian AmirAmjadi}
\author[5]{Parsa Hosseini}

\address[1]{Gilan University of Medical Sciences, Rasht, Iran}
\address[2]{University of Göttingen, Göttingen, Germany}
\address[3]{University of Tehran, Tehran, Iran}
\address[4]{Hamadan University of Technology, Hamadan, Iran}
\address[5]{Islamic Azad University, Tehran, Iran}

\begin{abstract}
The diagnosis of oral diseases presents a problematic clinical challenge, characterized by a wide spectrum of pathologies with overlapping symptomatology. To address this, we developed Clinical Semantic Intelligence (CSI), a novel artificial intelligence framework that diagnoses 118 different oral diseases by computationally modeling the cognitive processes of an expert clinician. Our core hypothesis is that moving beyond simple pattern matching to emulate expert reasoning is critical to building clinically useful diagnostic aids.

CSI's architecture integrates a fine-tuned multimodal CLIP model with a specialized ChatGLM-6B language model. This system executes a Hierarchical Diagnostic Reasoning Tree (HDRT), a structured framework that distills the systematic, multi-step logic of differential diagnosis. The framework operates in two modes: a Fast Mode for rapid screening and a Standard Mode that leverages the full HDRT for an interactive and in-depth diagnostic workup.

To train and validate our system, we curated a primary dataset of 4,310 images, supplemented by an external hold-out set of 176 images for final validation. A clinically-informed augmentation strategy expanded our training data to over 30,000 image-text pairs. On a 431-image internal test set, CSI's Fast Mode achieved an accuracy of 73.4\%, which increased to 89.5\% with the HDRT-driven Standard Mode. The performance gain is directly attributable to the hierarchical reasoning process. Herein, we detail the architectural philosophy, development, and rigorous evaluation of the CSI framework.

\end{abstract}

\begin{keyword}
Oral Disease Diagnosis \sep Multimodal AI \sep Clinical Decision Support \sep Contrastive Language Image Pre-training \sep Hierarchical Reasoning \sep ChatGLM-6B
\end{keyword}

\end{frontmatter}

\section{Introduction}

\subsection{The Clinical Challenge of Oral Diagnosis}
The accurate and timely diagnosis of oral diseases is a cornerstone of patient care, yet it remains a persistent clinical challenge. The vast range of pathologies, from common inflammatory conditions to rare malignancies, frequently present with ambiguous and overlapping clinical features \cite{youngblood2017,alrashdan2016,scully2022}. This ambiguity can lead to diagnostic uncertainty, resulting in significant delays and misdiagnoses. In fact, studies have shown that up to 14\% of patients with oral cancer experience such delays, which is directly correlated with poorer prognoses \cite{scott2022}. The diagnostic problem is further compounded by the relative rarity of certain conditions, such as pemphigus vulgaris or Kaposi sarcoma, which limits the clinical exposure of many practitioners \cite{akpan2021,cattelan2020}.

\subsection{Our Approach: Emulating the Mind of the Expert Clinician}
Conventional AI approaches to medical imaging often treat diagnosis as a classification task, which does not capture the nuanced, iterative reasoning process of an expert clinician. Our work is based on a different philosophy: To build a truly effective diagnostic aid, AI must emulate the cognitive framework of a specialist. To that end, we developed \textbf{CSI (Clinical Semantic Intelligence)}, a system designed to "think" like an oral pathologist.

Our design philosophy is rooted in modeling two key aspects of expert cognition.
\begin{enumerate}
    \item \textbf{Dual-Speed Reasoning:} Specialists fluidly switch between rapid, intuitive pattern recognition for classic cases and a slower, more deliberate analytical process for ambiguous presentations. CSI mirrors this with its \textit{Fast Mode} and \textit{Standard Mode}.
    \item \textbf{Hierarchical, Reductive Logic:} For a complex case, an expert does not simply guess. They systematically narrow the possibilities through a structured inquiry, a process that we computationally model with our \textit{Hierarchical Diagnostic Reasoning Tree (HDRT)}.
\end{enumerate}
By integrating multimodal data analysis with this expert-emulating reasoning structure, CSI aims to provide a diagnostic tool that is not only accurate but also transparent and clinically intuitive.

\subsection{Objectives}
The primary objective of this study was to develop and rigorously validate an AI model capable of accurately diagnosing a comprehensive range of 118 oral diseases. Our central hypothesis was that by computationally modeling the hierarchical reasoning process of clinicians, we could achieve a significant improvement in diagnostic precision over standard, non-interactive classification methods. We sought to build a system that could serve as a reliable decision support tool, with the ultimate goal of reducing diagnostic errors and improving patient oral health.

\begin{figure}[H]
    \centering
    \includegraphics[width=1.1\linewidth]{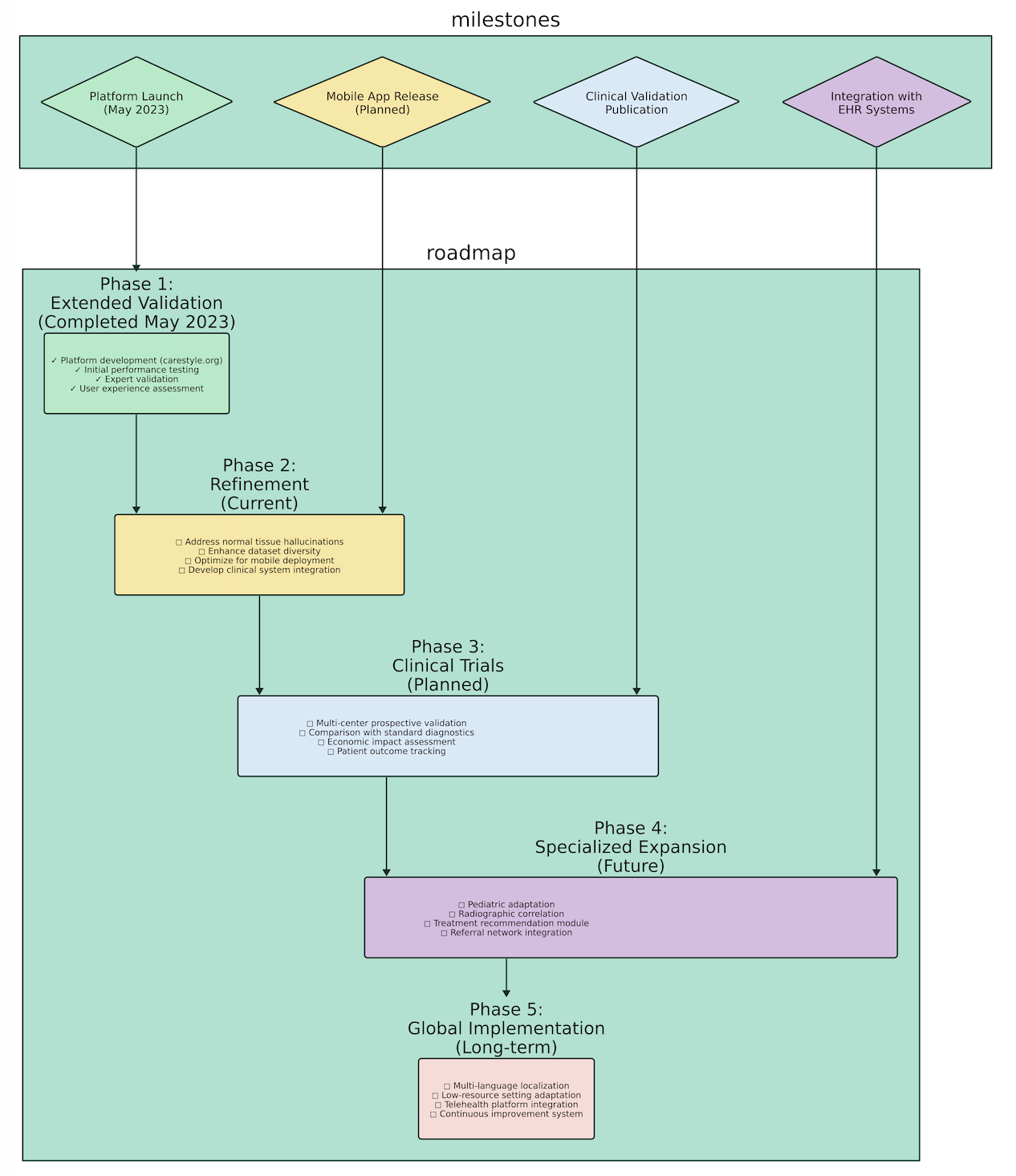}
    \caption{Project timeline from April 2023 to September 2023, depicting the major development milestones of the CSI framework.}
    \label{fig:Roadmap}
\end{figure}

\section{Literature Review and Related Work}
The challenge of differential diagnosis in oral pathology is well documented \cite{neville2015}. Although AI in dentistry has shown promise, particularly through the application of CNNs to radiographic analysis \cite{lee2018}, most systems address narrow tasks and lack the capacity for comprehensive and interactive differential diagnosis. The advent of powerful multimodal models like CLIP has shown the potential of aligning visual and textual data for a deeper semantic understanding \cite{radford2021}. Currently, the integration of large language models into interactive systems offers a pathway to resolve clinical ambiguity through discussions \cite{li2020}. Our work builds on these advances but addresses a critical gap by combining them with the principles of hierarchical decision modeling \cite{quinlan1986}, a concept long established in medical informatics. CSI's novelty lies in synthesizing these three parts, multimodal learning, interactive dialogue, and hierarchical reasoning, into a single cohesive framework that emulates the full arc of expert clinical evaluation.

\section{Dataset and Methods}
\subsection{Dataset Curation and Partitioning}
To build a robust training corpus, we curated a primary dataset of 4,310 high-quality clinical images. Archival images (dating from 2021 to 2023) were curated and augmented between April and September 2023, sourced from authoritative medical textbooks, collaborations with multiple dental clinics, and direct contributions from clinical practices (see Acknowledgments). To establish a reliable ground truth for evaluation, all images in our test set were verified on the basis of biopsy results, where applicable. The primary data set was partitioned for the development of the model, and a separate external validation set of 176 images was reserved for the final and unbiased comparison (Table \ref{tab:dataset_partition}).

\begin{table}[H]
  \centering
  \caption{Partitioning of the primary and external validation datasets.}
  \label{tab:dataset_partition}
  \begin{tabular}{@{}llll@{}}
    \toprule
    \textbf{Partition}        & \textbf{Percentage} & \textbf{Image Count} & \textbf{Purpose}                              \\
    \midrule
    Training             & 70\%                & 3,017         & Model training                                \\
    Validation            & 20\%                & 862           & Hyperparam. tuning                         \\
    Internal Test        & 10\%                & 431                 & Internal evaluations                        \\
    \midrule
    External Validation  & N/A                 & 176                 & Unbiased benchmark     \\
                         &                 &                  & on complex cases     \\
    \bottomrule
  \end{tabular}
\end{table}

\subsection{Data Augmentation Strategy}
To mitigate class imbalance inherent in clinical data, we developed a clinically-grounded augmentation strategy that expanded our dataset to approximately 30,000 image-text pairs. This involved both pathology-preserving image transformations and text-based augmentation to create diverse yet realistic clinical scenarios. To systematically balance the dataset, we employed a distribution-aware sampling formula:
\begin{equation}
\text{Augmentation Factor} = \log(b) \;+\; a \times \Bigl(1 - \tfrac{N_{\mathrm{initial}}}{N_{\max}}\Bigr)
\end{equation}
\(N_{\mathrm{initial}}\) is the image count for a given disease, and \(N_{\max}\) is the count for the most common disease. Based on empirical validation, where different settings were evaluated on a held-out portion of the training set to identify a configuration that provided a satisfactory balance, as confirmed by expert review, we set the tuning parameters at a = 10 and b = 2.

\subsection{Diagnostic Zones and The Expert-Corrected Atlas}
\begin{figure}[H]
    \centering
    \includegraphics[width=1.0\linewidth]{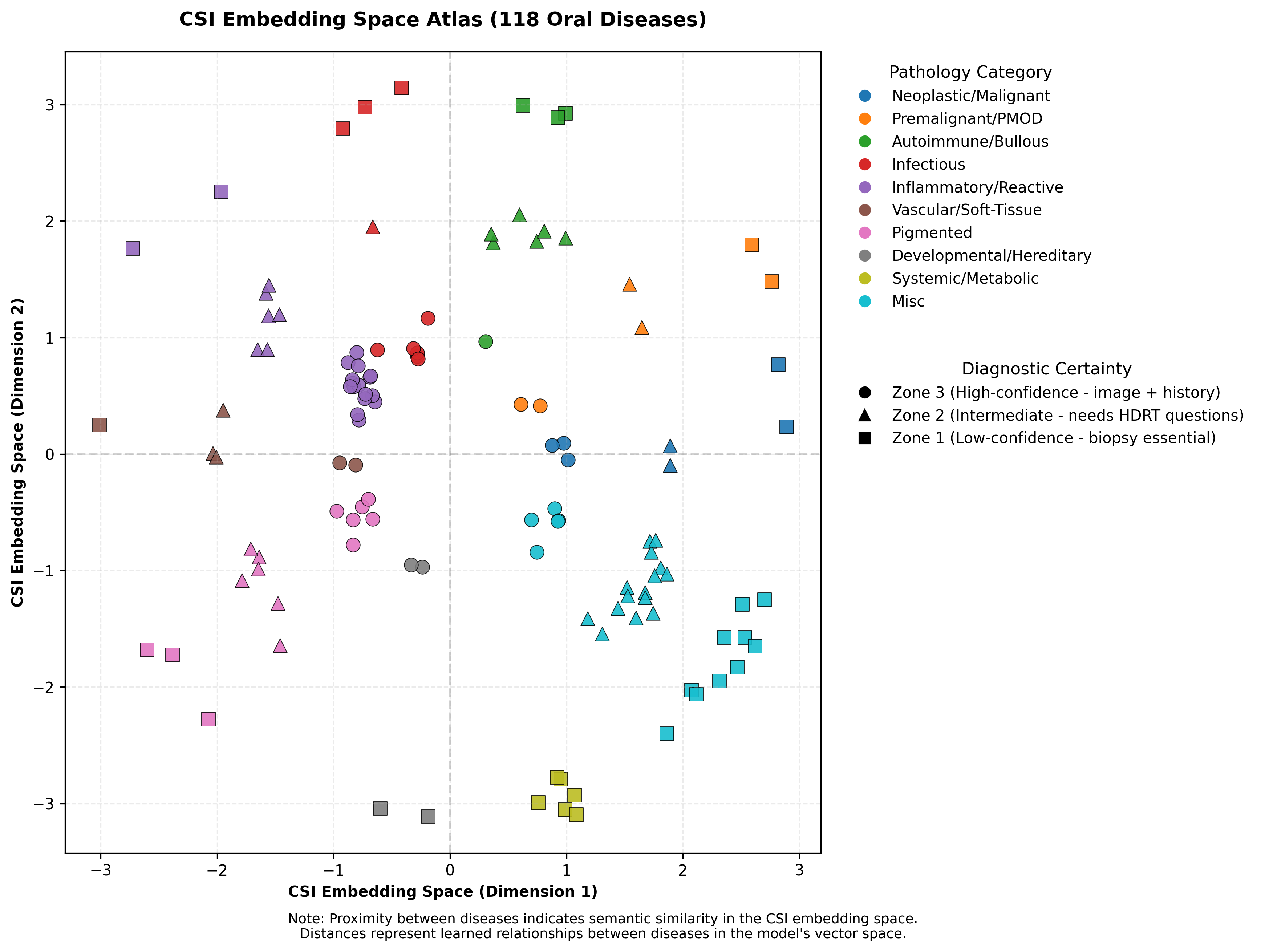}
    \caption{The CSI Embedding Space Atlas, mapping 118 oral diseases. Colors represent pathology families, shapes represent diagnostic zones (circle: Zone 3, triangle: Zone 2, square: Zone 1), and distance from the center correlates with diagnostic certainty.}
    \label{fig:atlas}
\end{figure}

Following training, we used t-SNE to visualize the model's learned embedding space for all 118 diseases. This visualization, which we termed the \textbf{Expert-Corrected Diagnostic Atlas} (Figure \ref{fig:atlas}), revealed how the model organizes diseases based on their semantic and visual features. From this analysis, and validated by a panel of five oral pathologists, we defined three diagnostic zones that categorize diseases by their intrinsic diagnostic difficulty:
\begin{itemize}
    \item \textbf{Zone 3 (Routine):} Conditions with pathognomonic features.
    \item \textbf{Zone 2 (Intermediate):} Conditions with significant overlap of characteristics requiring clinical context.
    \item \textbf{Zone 1 (Complex):} Rare or diagnostically challenging conditions that often require biopsy.
\end{itemize}
The internal test set (N = 431) consisted of 241 images of Zone 3 (55.9\%), 153 images of Zone 2 (35.5\%), and 37 images of Zone 1 (8.6\%).

\subsection{Model Architecture}
The architectural philosophy of CSI is rooted in an additive approach, enhancing powerful foundation models with specialized clinical knowledge without compromising their general capabilities.

\subsubsection{The Multimodal Fusion Model: Creating a Clinical Gestalt}
The diagnostic process begins with a fusion model that integrates visual and textual input to create a unified semantic representation: a computational analog of a clinician's initial "gestalt." We used a CLIP ViT-H/14 architecture, fine-tuned with the Wise-FT method \cite{wortsman2022} to prevent catastrophic forgetting. The image and text features are projected into a common embedding space and fused into a single 1024-dimensional vector that encapsulates the case.

\begin{figure}[H]
    \centering
    \includegraphics[width=1.05\linewidth]{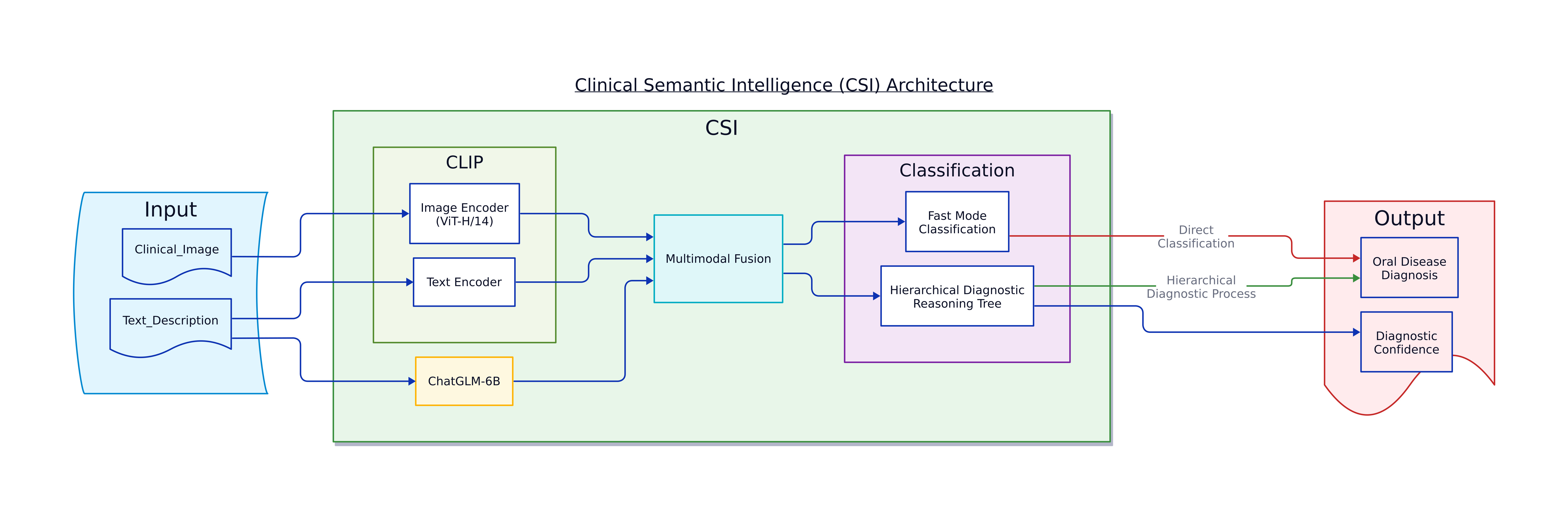}
    \caption{High-level overview of the CSI model architecture. It depicts the initial fusion of multimodal inputs to form a semantic representation, which is then processed by the dual-capability ChatGLM-6B model executing the HDRT.}
    \label{fig:model_arch}
\end{figure}

\subsubsection{ChatGLM-6B: A Dual-Capability Language Engine}
Our central innovation was to endow a single ChatGLM-6B model with two distinct capabilities through sequential supervised fine-tuning (SFT) on a 40 million token dental corpus.
\begin{enumerate}
    \item \textbf{First SFT (Clinical Communicator and Router):} This phase trained the model to be an intelligent clinical interface, capable of managing conversation and routing queries appropriately.
    \item \textbf{Second SFT (HDRT Executor):} This phase trained the same model to accept the semantic vector of the fusion model and execute the structured reasoning of the HDRT.
\end{enumerate}

\subsubsection{The Hierarchical Diagnostic Reasoning Tree (HDRT)}
The HDRT is the computational core of CSI's Standard Mode. It models the systematic, reductive reasoning of an expert by proceeding through six distinct levels of analysis. It is crucial to distinguish the six HDRT \textit{levels}, which describe the sequential steps in the reasoning process, from the three diagnostic \textit{zones}, which categorize the diseases themselves by difficulty. To handle uncertainty, a confidence threshold gating mechanism asks the user for more information if the top two diagnoses have similar probabilities, thereby simulating a clinical dialogue.

\begin{figure}[H]
    \centering
    \includegraphics[width=1.1\linewidth]{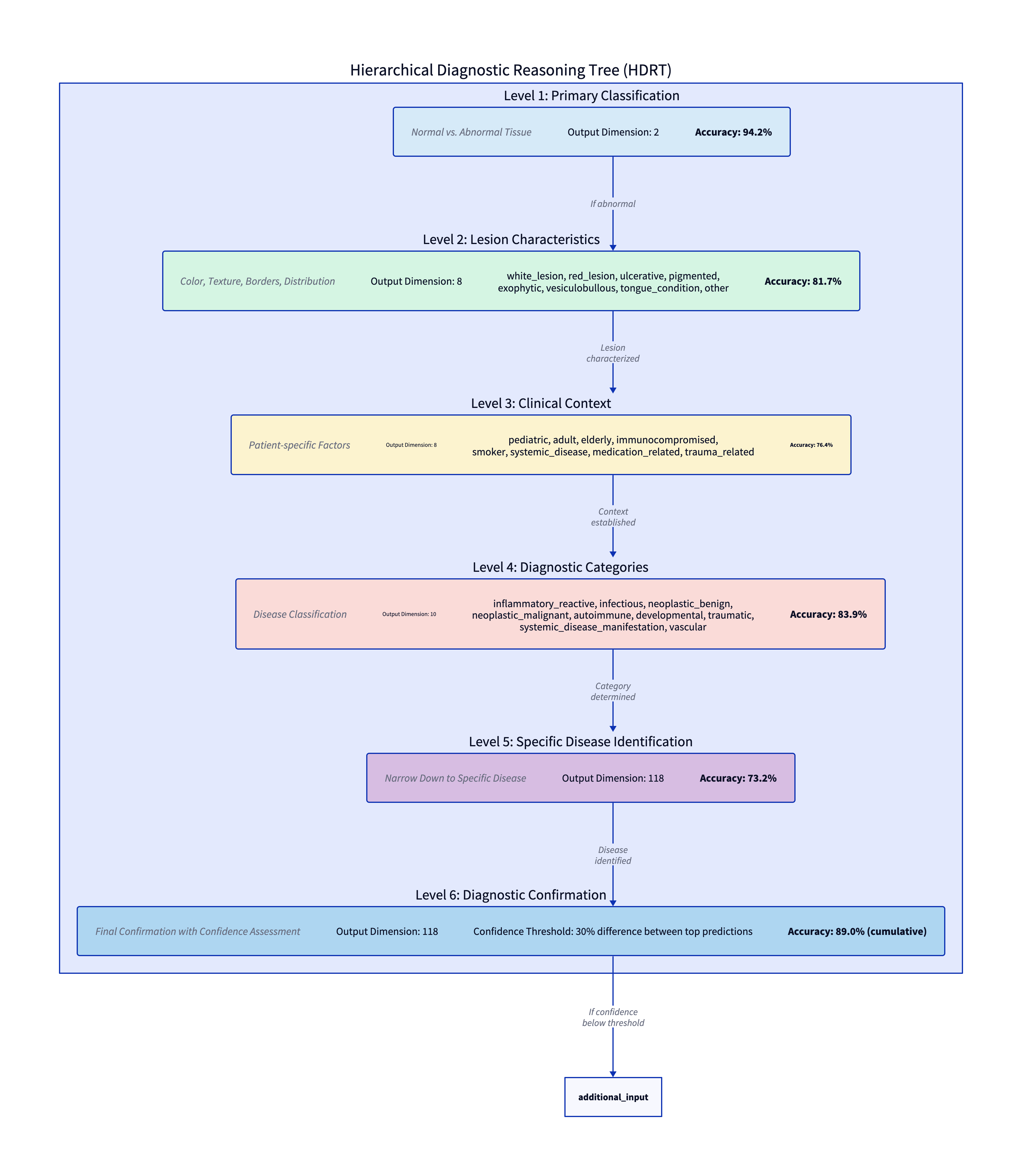} 
    \caption{Conceptual diagram of the Hierarchical Diagnostic Reasoning Tree (HDRT). This figure illustrates the high-level flow of the six-level clinical reasoning process. The specific operations and logic at each level are detailed in Appendix B.}
    \label{fig:hdrt_diagram}
\end{figure}

\section{Results}
\subsection{Performance on the Internal Test Set}
We first evaluated CSI's performance on the 431-image internal test set. The results, shown in Table \ref{tab:zone_accuracy}, highlight the substantial performance gain provided by the HDRT framework. For comparison, GPT-4 was evaluated in a non-interactive, zero-shot setting.

\begin{table}[H]
  \centering
  \small
  \begin{threeparttable}
  \caption{Zone-specific accuracy (\%) on the internal test set (N=431).}
  \label{tab:zone_accuracy}
    \begin{tabular*}{\textwidth}{@{\extracolsep{\fill}}lcccc}
    \toprule
    \textbf{Zone} & \textbf{Test Images} & \textbf{Fast Mode} & \textbf{Standard Mode} & \textbf{GPT-4} \\
    \midrule
    Zone 3 (Routine)       & 241 & 85.4 & 94.3 & 79.4 \\
    Zone 2 (Intermediate)  & 153 & 63.7 & 89.5 & 60.2 \\
    Zone 1 (Complex)       & 37  & 35.8 & 58.1 & 38.7 \\
    \midrule
    \textbf{Overall}    & \textbf{431} & \textbf{73.4} & \textbf{89.5} & \textbf{69.1} \\
    \bottomrule
  \end{tabular*}
  \begin{tablenotes}
    \item[*] Overall accuracy is the weighted average calculated based on the number of test images in each diagnostic zone.
  \end{tablenotes}
  \end{threeparttable}
\end{table}

The greatest performance differential was observed in Zone 2, where HDRT improved accuracy by 25.8 percentage points, underscoring the value of contextual reasoning for diseases with overlapping features.

\subsection{Performance on the External Validation Dataset}
To assess the model's real-world generalization, we curated a challenging external validation set of 176 cases. This set was intentionally skewed towards more difficult cases, with a distribution of: Zone 1 (N=15, 8.5\%), Zone 2 (N=75, 42.6\%), and Zone 3 (N=86, 48.9\%).

\begin{table}[H]
\centering
\small
\begin{threeparttable}
\caption{Comparative performance (\%) on the complex external validation set (N=176).}
\label{tab:external_validation}
\begin{tabular*}{\textwidth}{@{\extracolsep{\fill}}lccc}
\toprule
\textbf{Metric} & \textbf{Fast Mode} & \textbf{Standard Mode} & \textbf{GPT-4} \\
\midrule
Zone 3 Accuracy & 77.9 & 90.7 & 68.6 \\
Zone 2 Accuracy & 56.0 & 85.3 & 57.3 \\
Zone 1 Accuracy & 33.3 & 53.3 & 13.3 \\
\midrule
\textbf{Overall Accuracy} & \textbf{64.8} & \textbf{85.2} & \textbf{58.5} \\
\bottomrule
\end{tabular*}
\begin{tablenotes}
    \item[*] Overall accuracy is the weighted average based on the zone distribution of this external set.
\end{tablenotes}
\end{threeparttable}
\end{table}

Even on this difficult, unseen dataset, CSI's Standard Mode demonstrated robust performance (85.2\%), maintaining a significant advantage over both its non-interactive counterpart and the generalist GPT-4 model.

\section{Discussion}
\subsection{The Value of Emulating Clinical Reasoning}
The central contribution of this work is a paradigm shift from conventional AI classification to the emulation of domain-specific diagnostic reasoning. By computationally modeling the hierarchical logic of an expert clinician, CSI achieves a "cognitive fidelity" that allows it to systematically navigate diagnostic uncertainty. This is not merely an academic exercise; it directly addresses the clinical challenge of differentiating pathologies with overlapping features. The superior performance of the Standard Mode over the Fast Mode provides clear evidence that, for complex medical tasks, the reasoning process itself is as important as the final classification.

\subsection{Clinical Implications and Comparative Performance}
The CSI framework has significant potential as both a clinical decision support tool and an educational platform. For practitioners in underserved areas, it can provide access to specialist-level reasoning. For trainees, the transparent step-by-step diagnostic pathways can serve as a powerful learning aid. Our findings indicate a clear performance advantage for the hierarchical specialized approach of CSI compared to the zero-shot capabilities of a generalist model like GPT-4 (85.2\% vs. 58.5\% on the external set).

However, a critical caveat is that this represents an asymmetric benchmark. We compared our interactive, multi-turn Standard Mode against a single-turn evaluation of GPT-4. This design effectively highlights the value added by the HDRT's interactive framework, but a direct comparison of conversational diagnostic capabilities would require a symmetric evaluation where the baseline model is also engaged in a structured dialogue. This benchmark was outside the scope of this study, but it represents a critical direction for future work.

\subsection{Limitations and Future Directions}
This study has several limitations. Our model can occasionally misclassify normal anatomical variants. Although our data set is extensive, it may not capture the full spectrum of global diversity in disease presentation. Furthermore, the performance of the interactive Standard Mode is inherently dependent on the quality of user-provided information. Future work will focus on expanding the dataset, optimizing the model for deployment in resource-limited settings, and enhancing the natural language interface. Ultimately, large-scale, multicenter clinical trials are required to definitively validate the real-world efficacy of the CSI framework.

\section{Conclusion}
Clinical Semantic Intelligence (CSI) provides a robust framework for AI-assisted oral diagnosis, distinguished by its foundational philosophy of emulating expert clinical reasoning. By integrating multimodal analysis with a hierarchical interactive diagnostic process, CSI successfully diagnoses a comprehensive range of 118 oral diseases with a high degree of precision and interpretability. Our results, particularly the 85.2\% accuracy achieved on a challenging external validation set, confirm that for specialized medical domains, an architecture that models the expert's cognitive process significantly outperforms generic, non-interactive approaches. Although further validation is necessary, the CSI framework represents a promising step toward developing AI systems that function not as black-box classifiers but as transparent and collaborative partners in the complex art of clinical diagnosis.

\section*{Acknowledgements}
The authors thank Dr. Babak Mashayekhi for his invaluable contribution of clinical images from his practice, which were essential for the curation of our datasets. The authors also acknowledge the support of the Gilan University of Medical Sciences in facilitating this research.

\newpage
\appendix
\section*{Appendix A: Comprehensive List of Oral Diseases Detectable by CSI}
{\small
\begin{enumerate}[label=\arabic*.]
    \item Pemphigus Vulgaris
    \item Lichenoid Contact Reaction
    \item Geographic Tongue
    \item Erythema Multiforme
    \item Acquired Hemangioma
    \item Verruciform Xanthoma
    \item Hemangioma
    \item Palatal Keratosis Associated with Reverse Smoking
    \item Denture Stomatitis
    \item Hereditary Benign Intraepithelial Dyskeratosis
    \item Necrotizing Ulcerative Gingivitis
    \item Squamous Cell Carcinoma
    \item Capillary Hemangioma
    \item Hypoadrenocorticism
    \item Recurrent Herpes Stomatitis
    \item Lichenoid Reaction
    \item Submucous Fibrosis
    \item Cytomegalovirus-Associated Ulceration
    \item Non-Hodgkin Lymphoma
    \item Verruca Vulgaris
    \item Schwannoma
    \item Lymphangioma
    \item Plasma Cell Stomatitis
    \item Pyogenic Granuloma
    \item Inflammation-Associated Hyperpigmentation
    \item Traumatic Ulcer
    \item Neurofibroma
    \item Smoker's Melanosis
    \item Angular Cheilitis
    \item Erythematous Lichen Planus
    \item Irritation Fibroma
    \item Physiologic Pigmentation
    \item Actinic Cheilitis
    \item Varicosities
    \item Mucocele and Ranula
    \item Herpangina
    \item Hyperthyroidism
    \item Pulp Polyp
    \item Melanotic Macule
    \item Candidal Leukoplakia
    \item Cushing's Disease
    \item Nicotinic Stomatitis
    \item Superficial Oral Burn
    \item Graft vs. Host Disease
    \item Epidermolysis Bullosa
    \item Lupus Erythematosus
    \item Peripheral Ossifying Fibroma
    \item Oral Leukoplakia
    \item Peripheral Giant Cell Granuloma
    \item Adenoid Cystic Carcinoma
    \item Drug-Induced Lichenoid Reaction
    \item Chemotherapy-Related Ulcers
    \item Erythroplakia
    \item Squamous Papilloma
    \item Giant Cell Fibroma
    \item Oral Lichen Planus
    \item Malignant Melanoma
    \item Verrucous Hyperplasia
    \item Pregnancy Tumor
    \item Eosinophilic Ulcer
    \item Vascular Malformation
    \item Vitamin B12 Deficiency
    \item Medicinal Metal-Induced Pigmentation
    \item Epulis Granulomatosum
    \item Smokeless Tobacco Keratosis
    \item Oral Hypersensitivity Reactions
    \item White Sponge Nevus
    \item Frictional Keratosis
    \item Hematoma-Ecchymosis-Petechia-Purpura
    \item Keratoacanthoma
    \item Laugier-Hunziker Pigmentation
    \item Lipoma
    \item Morsicatio
    \item Hereditary Hemorrhagic Telangiectasia
    \item Oral Hairy Leukoplakia
    \item Bullous Pemphigoid
    \item Epulis Fissuratum
    \item Parulis
    \item Freckle-Ephelis
    \item Verrucous Carcinoma
    \item Peutz-Jeghers Syndrome
    \item Melanocytic Nevus
    \item Leaflike Fibroma
    \item Ulcerative Squamous Cell Carcinoma
    \item Xeroderma Pigmentosum
    \item Amalgam Tattoo
    \item Vesiculobullous Diseases
    \item Necrotizing Sialometaplasia
    \item Recurrent Aphthous Stomatitis
    \item Oral Melanoacanthoma
    \item Graft-versus-Host Disease
    \item Reticular Lichen Planus
    \item Deep Fungal Ulceration
    \item Inflammatory Papillary Hyperplasia
    \item Kaposi Sarcoma
    \item Behçet's Disease
    \item Dyskeratosis Congenita
    \item Median Rhomboid Glossitis
    \item Erythematous Candidiasis
    \item Tuberculous Ulcer
    \item Mucous Membrane Pemphigoid
    \item Multifocal Epithelial Hyperplasia
    \item Chancre or Syphilitic Ulceration
    \item Leukoedema
    \item Shingles
    \item Café au Lait Pigmentation
    \item Cheilitis Glandularis
    \item Melasma
    \item Primary Herpetic Gingivostomatitis
    \item Cyclic Neutropenia
    \item Chronic Mucocutaneous Candidiasis
    \item Angiosarcoma
    \item Hairy Tongue
    \item Sustained Traumatic Ulcer
    \item Drug-Induced Melanosis
    \item Proliferative Verrucous Leukoplakia
    \item Pseudomembranous Candidiasis
    \item HIV-Associated Pigmentation
\end{enumerate}
}

\newpage

\section*{Appendix B: HDRT Framework Technical Details}
\subsection*{B.1 Multimodal Integration Details}
\begin{itemize}
    \item \textbf{Projection Layers:} 
    \begin{enumerate}
        \item CLIP Image Projection: Linear(1280$\to$1024)
        \item CLIP Text Projection: Linear(1024$\to$1024)
        \item GLM Projection: Linear(4096$\to$1024)
    \end{enumerate}
    \item \textbf{CLIP Fusion:} Concatenate image \& text features, reduce dimension (2048$\to$1024).
    \item \textbf{Multimodal Fusion:} Concatenate CLIP-fused \& GLM features (2048$\to$1024).
\end{itemize}

\subsection*{B.2 HDRT Level-Specific Dimensions}
\begin{itemize}
    \item \textbf{Level 1:} Primary classification (2 outputs: normal/abnormal).
    \begin{figure}[H]
        \centering
        \includegraphics[width=1.1\linewidth]{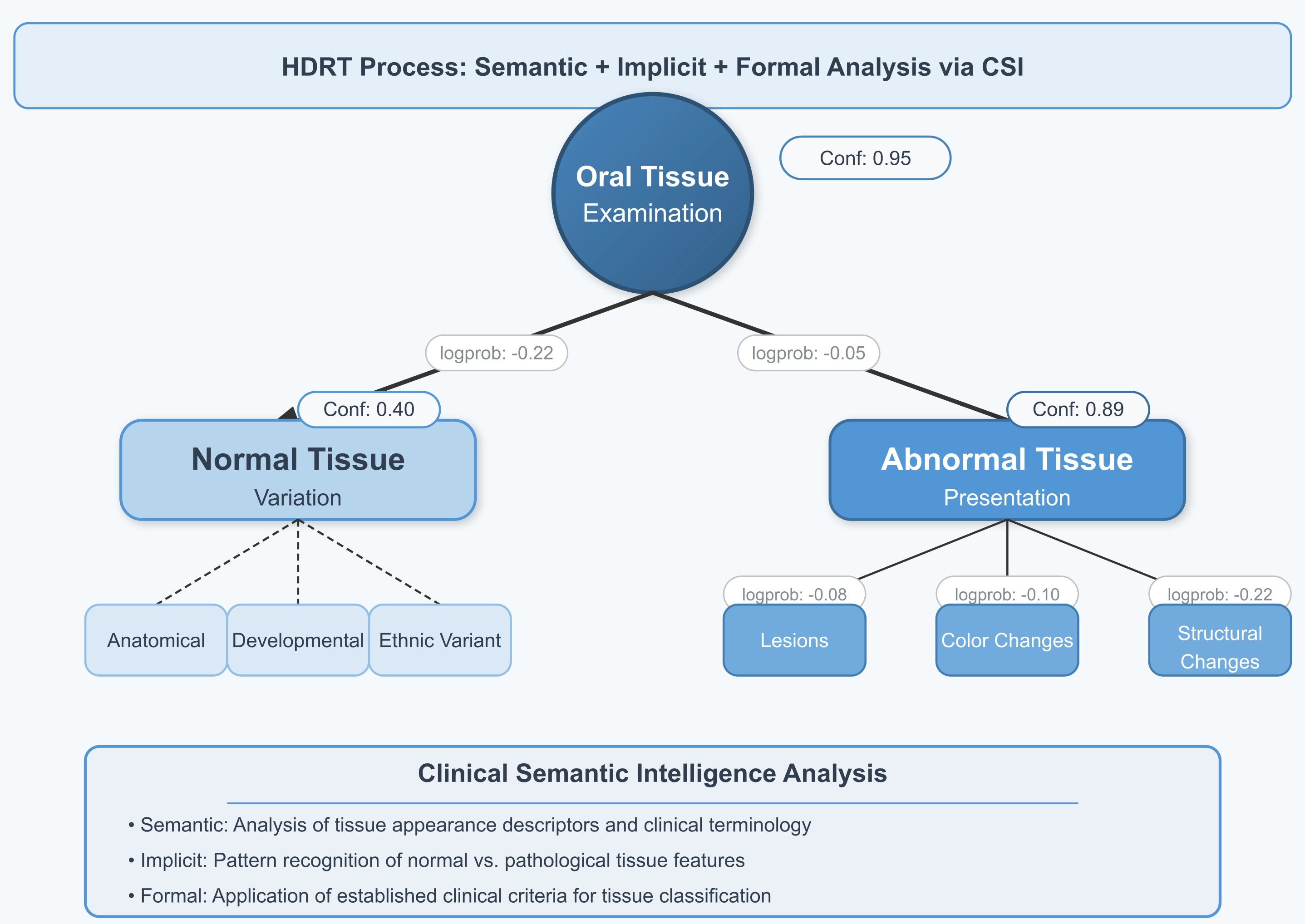} 
        \caption{Primary Classification: Initial assessment of oral tissue with confidence scoring}
        \label{fig:hdrt}
    \end{figure}

    \newpage

    \item \textbf{Level 2:} Lesion characteristics (8 outputs by color/texture).
    \begin{figure}[H]
        \centering
        \includegraphics[width=1.1\linewidth]{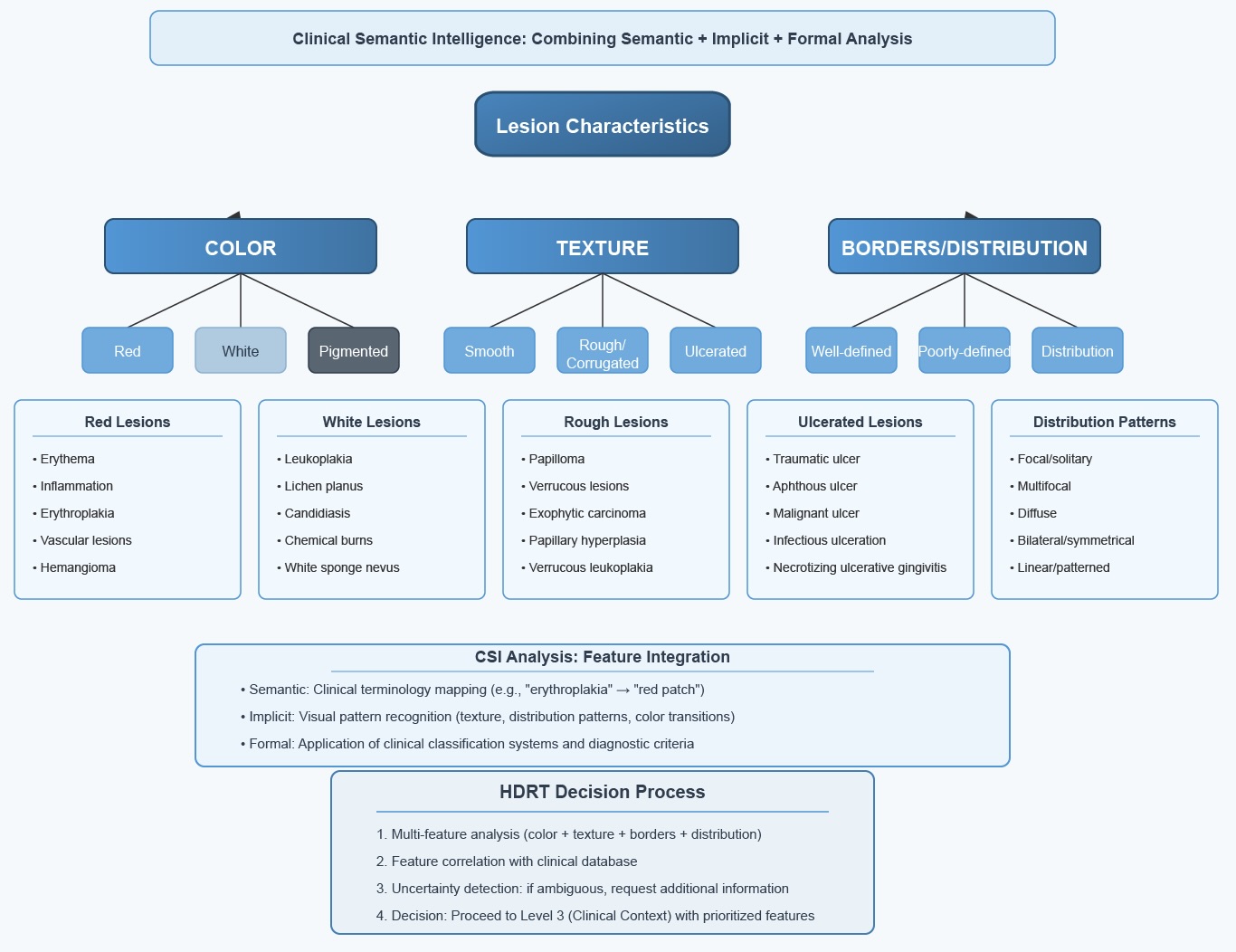} 
        \caption{Lesion Characteristics: Assessment of visible abnormalities with CSI analysis}
        \label{fig:hdrt}
    \end{figure}

    \newpage

    \item \textbf{Level 3:} Clinical context (8 outputs according to demographic/ risk).
    \begin{figure}[H]
        \centering
        \includegraphics[width=1.1\linewidth]{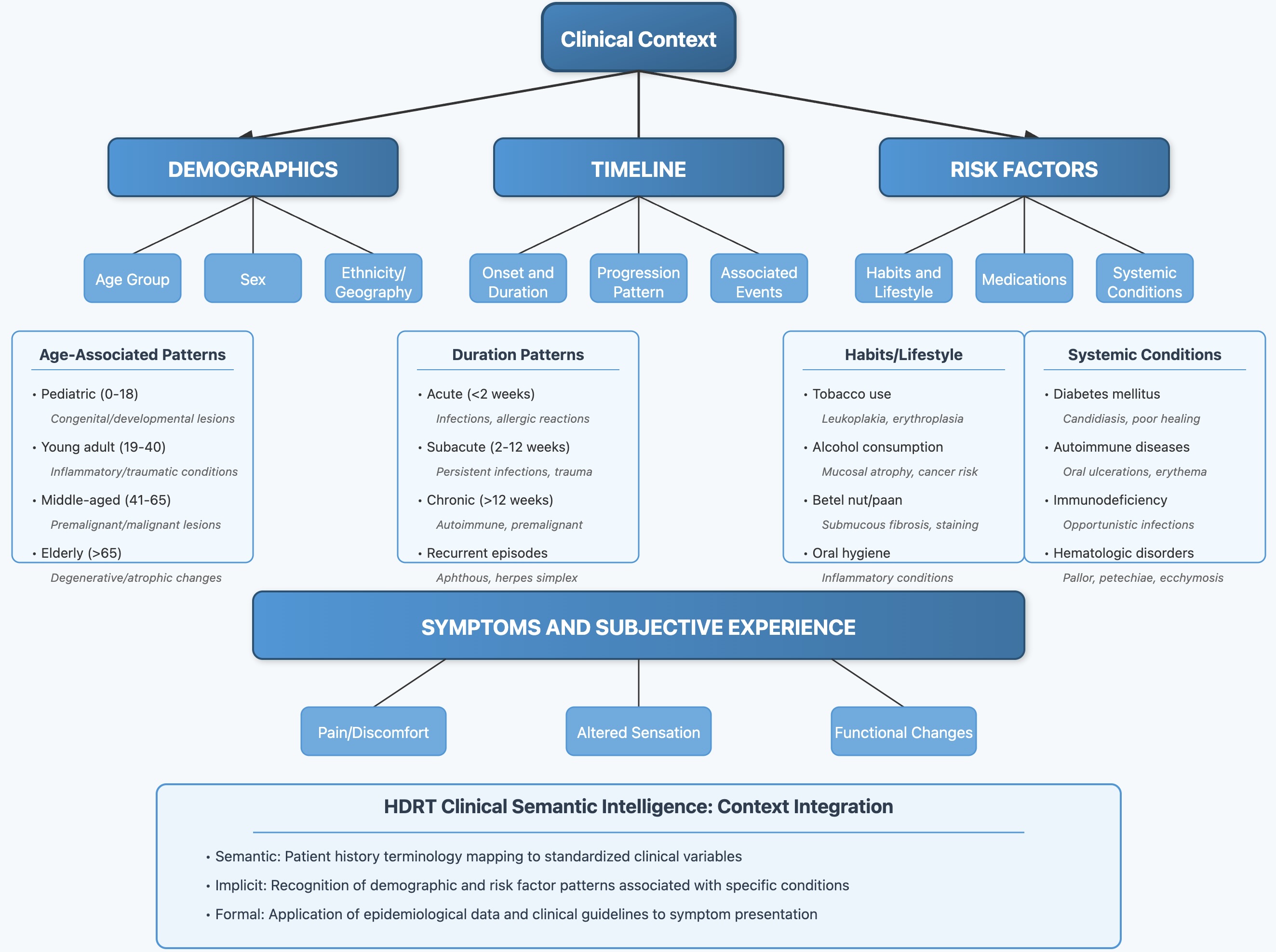} 
        \caption{Clinical Context: Patient-specific factors that influence diagnostic assessment.}
        \label{fig:hdrt}
    \end{figure}

    \newpage

    \item \textbf{Level 4:} Diagnostic categories (10 outputs: inflammatory, neoplastic, etc.).
    \begin{figure}[H]
        \centering
        \includegraphics[width=1.1\linewidth]{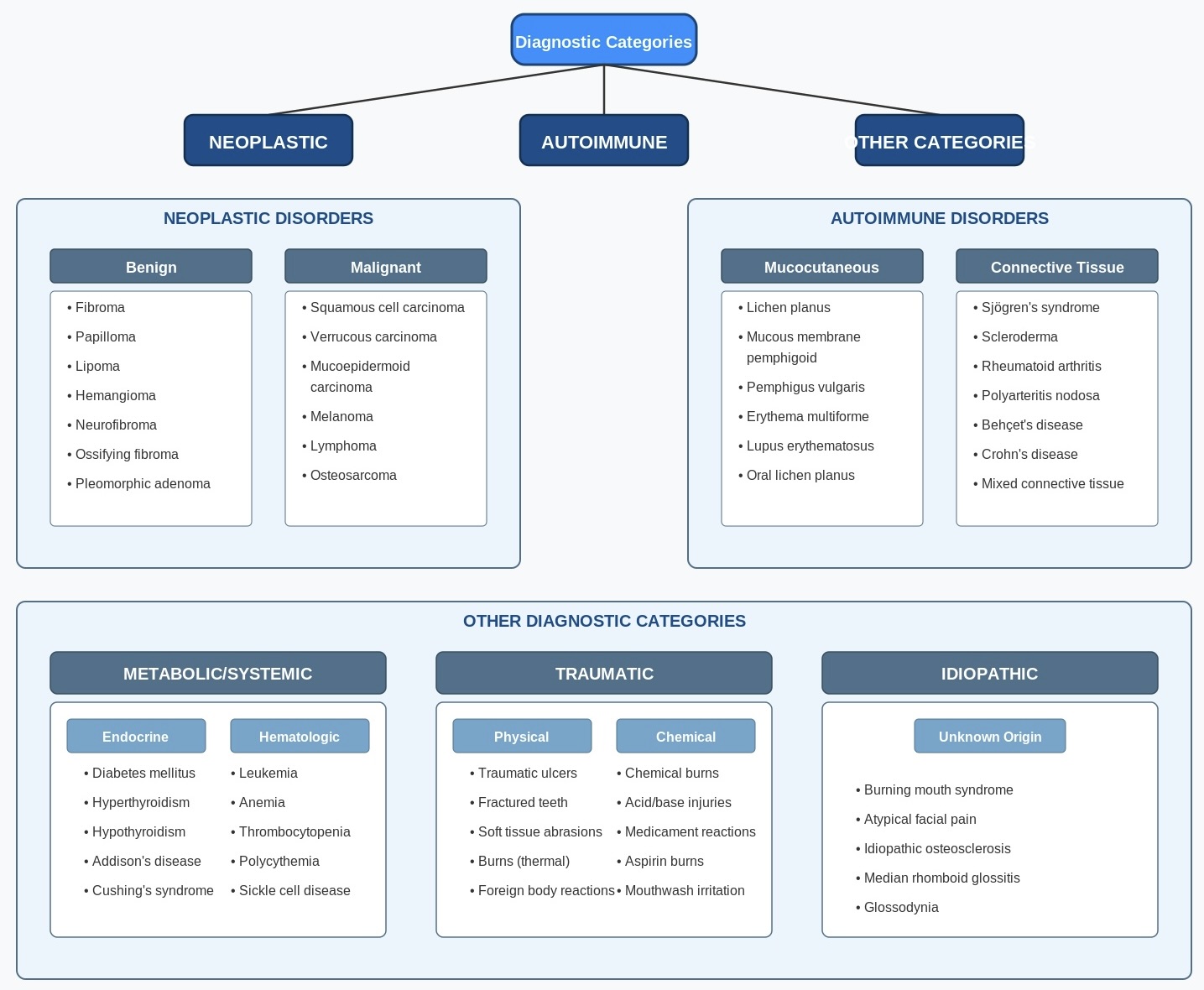} 
        \caption{Diagnostic Categories: Neoplastic, autoimmune, and additional groupings.}
        \label{fig:hdrt}
    \end{figure}

    \newpage

    \item \textbf{Level 5:} Disease identification (118 possible diseases).
    \begin{figure}[H]
        \centering
        \includegraphics[width=1.1\linewidth]{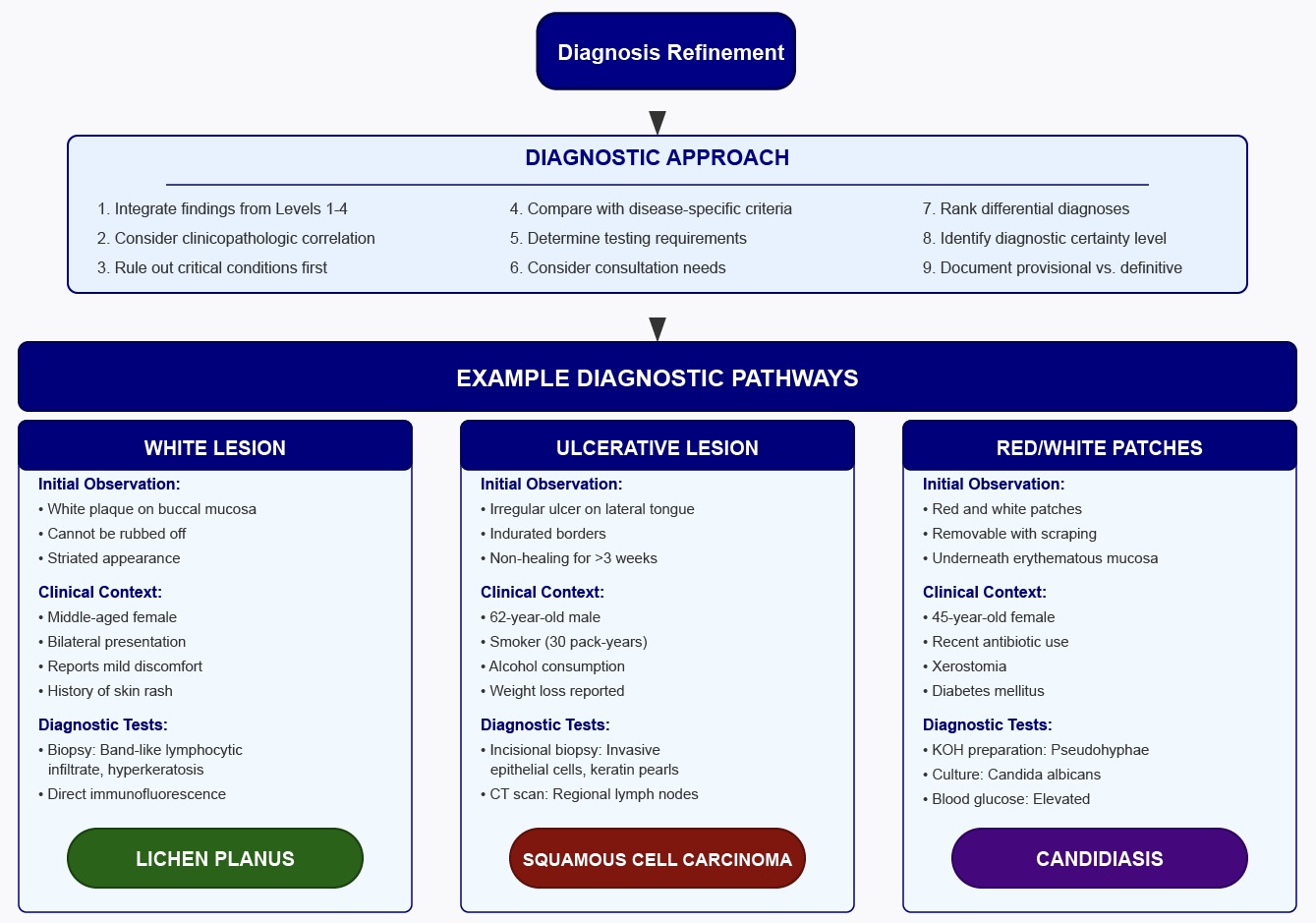} 
        \caption{Specific Disease Identification: Narrowing down to a specific diagnosis based on combined evidence (covering 118 possible conditions).}
        \label{fig:hdrt}
    \end{figure}

    \newpage

    \item \textbf{Level 6:} Diagnostic confirmation (118 final probabilities).
    \begin{figure}[H]
        \centering
        \includegraphics[width=1.1\linewidth]{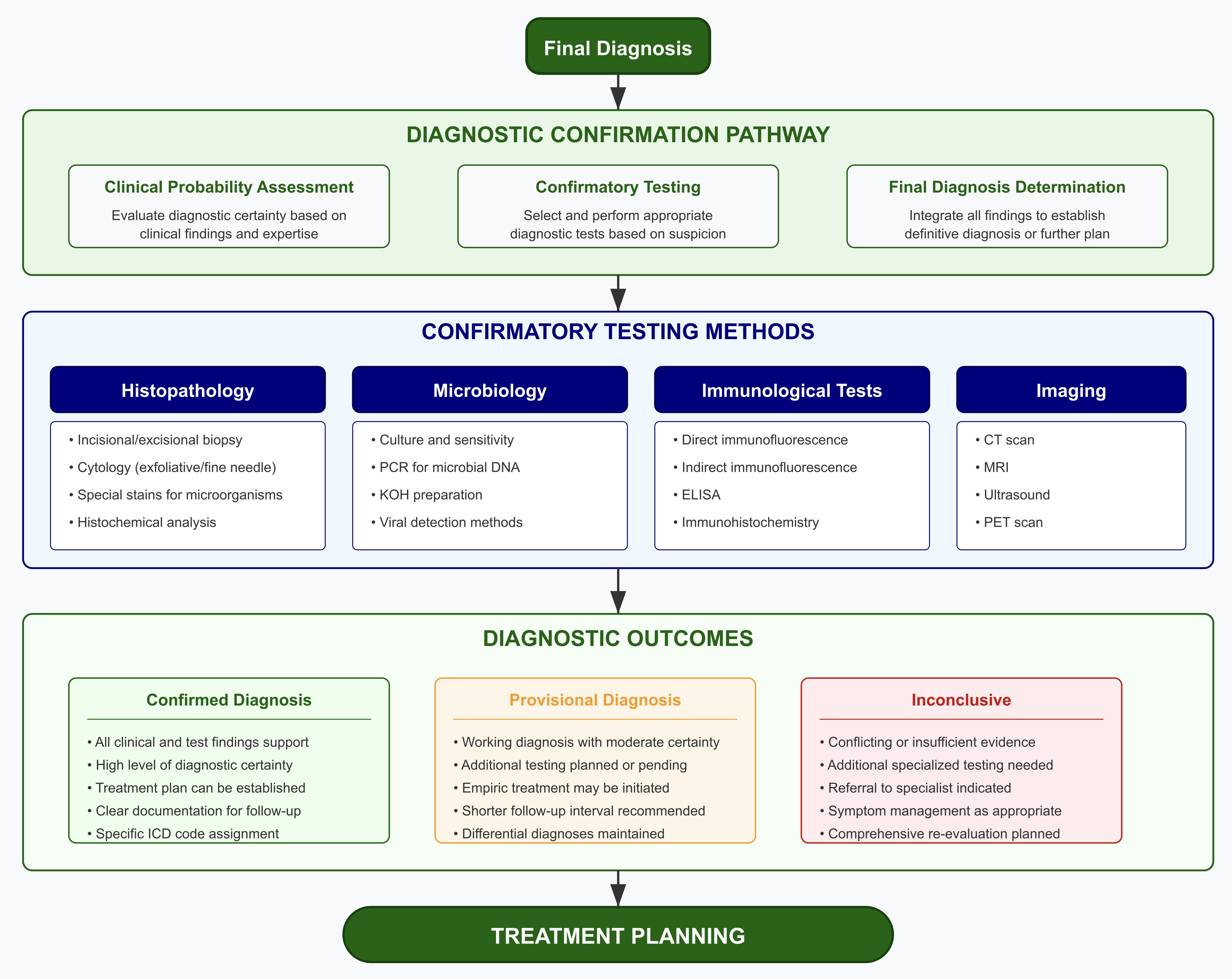} 
        \caption{Diagnostic Confirmation: Final assessment, confirmatory testing, and treatment planning.}
        \label{fig:hdrt}
    \end{figure}

\end{itemize}

\subsection*{B.3 Confidence-Threshold Gating Implementation}
When the difference between the logarithmic probabilities of the top two ranked diagnoses is below a threshold of 0.3, the system flags the uncertainty and is prompted to request additional clarification input from the user.

\newpage

\section*{Appendix C: CSI Output Examples}
\subsection*{C.1 Fast Mode Output Example}
\begin{itemize}
    \item \textbf{Primary Diagnosis:} Oral Lichen Planus (78.5\% confidence)
    \item \textbf{Differential Diagnosis:}\\
    Lichenoid Reaction (12.3\%)\\
    Leukoplakia (5.7\%)\\
    Candidal Leukoplakia (2.1\%)
\end{itemize}

\subsection*{C.2 Standard Mode Output Example}
\begin{itemize}
    \item \textbf{Diagnostic Path:} Abnormal $\to$ White Lesion $\to$ Adult $\to$ Inflammatory/Reactive $\to$ Oral Lichen Planus
    \item \textbf{Final Diagnosis:} Oral Lichen Planus (85.3\% confidence, high certainty)
    \item \textbf{Differential Diagnosis:} 1. Lichenoid reaction (8.1\%), 2. Leukoplakia (3.4\%)
    \item \textbf{Diagnostic Conversation:}
    \begin{enumerate}
        \item \textit{Q1:} What are the main visual characteristics of this lesion?
        \item \textit{A1:} White reticular pattern in the buccal mucosa, bilateral presentation.
        \item \textit{Q2:} Any significant history of medications?
        \item \textit{A2:} No recent changes, no known triggers.
    \end{enumerate}
\end{itemize}

\subsection*{C.3 ChatGLM-Enhanced Interactive Response Example}
\textit{User Query:} "What is the difference between oral lichen planus and that lichen-like reaction?"
\newline
\textbf{CSI Response:}
\begin{quote}
Based on the diagnostic assessment, several key differences can be highlighted between Oral Lichen Planus and a Lichenoid Reaction:
\begin{enumerate}
    \item Bilateral, reticular lesions with characteristic Wickham's striae strongly favor Oral Lichen Planus, whereas Lichenoid Reactions tend to be more localized or unilateral.
    \item The absence of known triggers, such as new medications or dental materials, reduces the likelihood of a lichenoid reaction.
    \item A mild burning sensation is more typical for Oral Lichen Planus.
\end{enumerate}
Based on this reasoning, confidence is 85.3\% for Oral Lichen Planus versus 8.1\% for a Lichenoid Reaction. A biopsy may be recommended for definitive differentiation.
\end{quote}

\end{document}